\documentclass[runningheads]{llncs}
\usepackage{xcolor}
\usepackage{graphicx}
\usepackage{subfigure}
\usepackage{times}
\usepackage{epsfig}
\usepackage{graphicx}
\usepackage{amsmath}
\usepackage{amssymb}
\usepackage{xfrac}
\usepackage{comment}
\usepackage{amsmath,amssymb}
\usepackage{tabularx}
\usepackage{rotating}
\usepackage{amsfonts}
\usepackage{booktabs}
\usepackage{siunitx}

\usepackage{multirow}
\usepackage{multirow}
\usepackage{xspace}

\PassOptionsToPackage{warn}{textcomp}  %
\usepackage{textcomp}                  %
\usepackage{mathptmx}                  %
\usepackage{times}                     %
\usepackage{cite}                      %
\usepackage{tabu}                      %
\usepackage{booktabs}                  %
\usepackage{mathrsfs}  
\usepackage{wrapfig}
\def\ie{\textit{i.e.},\xspace}

\def\eg{\textit{e.g.},\xspace}
\def\etal{\textit{et al}.\xspace}

\def\bfE{{\bf E}}

\def\bfS{{\bf S}}
\def\bfs{{\bf s}}

\def\bfB{{\bf B}}

\def\bfE{{\bf E}}
\def\bfp{{\bf p}}

\def\bfK{{\bf K}}

\def\bfT{{\bf T}}

\def\so3{{ \mathfrak{so}(3)}}
\def\SO3{{ \text{SO(3)}}}

\def\so3{{ \mathfrak{so}(3)}}
\def\se3{{ \mathfrak{se}(3)}}
\begin{document}
\title{HDM-Net: Monocular Non-Rigid 3D Reconstruction \\with Learned Deformation Model}
\titlerunning{HDM-Net: Monocular Non-Rigid 3D Reconstruction}
\author{Vladislav Golyanik$^{\flat,\sharp}$\and
Soshi Shimada$^{\flat,\sharp}$ \and
Kiran Varanasi$^\flat$\and
Didier Stricker$^{\flat,\sharp}$}
\authorrunning{V. Golyanik \etal}
\institute{
$^\flat$Augmented Vision, DFKI (\url{https://av.dfki.de}) $\;\;$ $^\sharp$University of Kaiserslautern %
}
\maketitle              %
\begin{abstract}
Monocular dense 3D reconstruction of deformable objects is a hard ill-posed problem in computer vision. 
Current techniques either require dense correspondences and rely on motion and deformation cues, or assume a highly accurate 
reconstruction (referred to as a template) of at least a single frame given in advance and operate in the manner of non-rigid tracking. 
Accurate computation of dense point tracks often requires multiple frames and might be computationally expensive.
Availability of a template is a very strong prior which restricts system operation to a pre-defined environment and scenarios. 
In this work, we propose a new hybrid approach for monocular non-rigid reconstruction which we call %
\textit{Hybrid Deformation Model Network} (HDM-Net). 
In our approach, a deformation model is learned by a deep neural network, with a combination 
of domain-specific loss functions. We train the network with multiple states of a non-rigidly deforming structure with a known shape 
at rest. HDM-Net learns different reconstruction cues including texture-dependent surface deformations, 
shading and contours. 
We show generalisability of HDM-Net to states not presented in the training dataset, with unseen textures and under 
new illumination conditions. 
Experiments with noisy data and a comparison with other methods demonstrate the robustness and accuracy of the proposed approach 
and suggest possible application scenarios of the new technique in interventional diagnostics and augmented reality.

\begin{figure} 
 \centering 
  \includegraphics[width=0.9\linewidth]{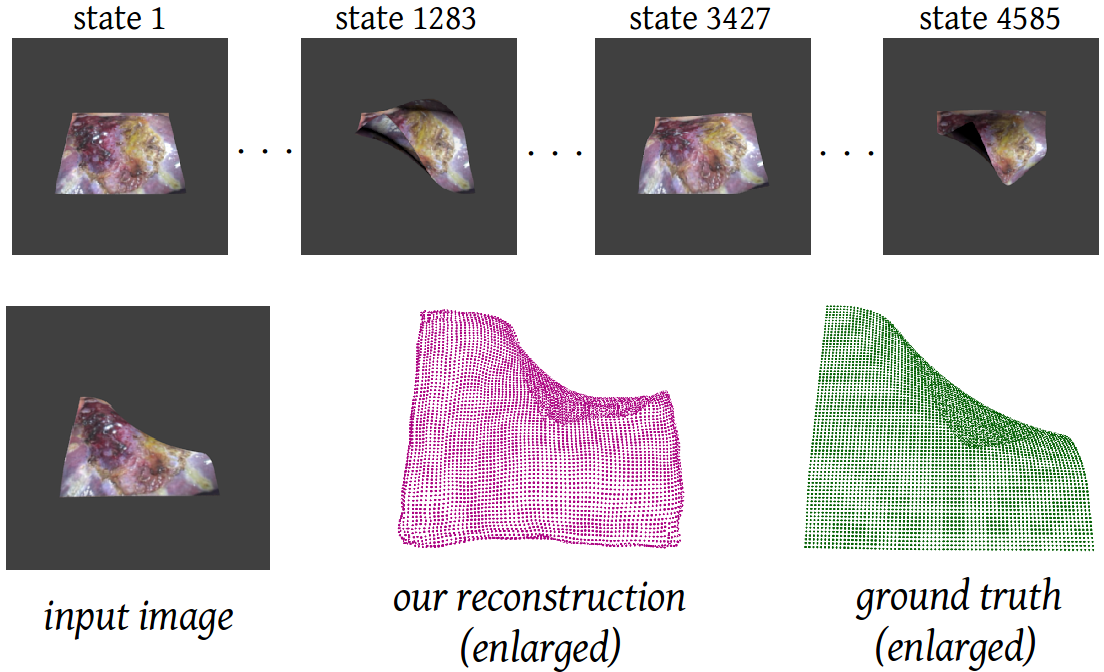} 
  \caption{Reconstruction of an endoscopically textured surface with the proposed HDM-Net. The network is trained on a textured synthetic 
  image sequence with ground truth geometry and accurately reconstructs unseen views in a small fraction of a second (\texttildelow 5ms).
  Our architecture is potentially suitable for real-time augmented reality applications. } 
  \label{06_data_set_and_comparison} 
\end{figure} 

\end{abstract}
\section{Introduction}

The objective of monocular non-rigid 3D reconstruction (MNR) is the recovery of a time-varying geometry observed 
by a single moving camera. In the general case, none of the states is observed from multiple views, 
and at the same time, both the object and the camera move rigidly. This problem is highly ill-posed in the sense 
of Hadamard since multiple states can cause similar 2D observations. To obtain a reasonable solution, multiple 
additional priors about the scene, types of motions and deformations as well as camera trajectory are required. 
Application domains of MNR are numerous and include robotics, medical applications and visual communication systems. 
MNR also has a long history in augmented reality (AR), and multiple applications 
have been proposed over the last twenty years ranging from medical systems to communication and entertainment \cite{Cohens1992, Haouchine2014}. 

All approaches to MNR can be divided into two main model-based classes --- non-rigid structure from motion (NRSfM) and template-based 
reconstruction (TBR). NRSfM relies on motion and deformation cues and requires dense point correspondences over multiple frames 
\cite{Garg2013, Golyanik2017}. Most accurate methods for dense correspondences operate on multiple frames and are prohibitively slow 
for real-time applications \cite{Taetz2016}. Moreover, their accuracy is volatile and influenced by changing illumination and shading effects 
in the scene. TBR, per definition, assumes a known template of the scene or an object, \ie a highly accurate reconstruction 
for at least one frame of the scene \cite{Perriollat2011, Yu2015}. Sometimes, the template also needs to be accurately positioned, with a minimal initial 
reprojection error to the reference frame. In this context, TBR can also be comprehended as non-rigid tracking \cite{Salzmann2007A}. 
Obtaining a template is beyond the scope of TBR, though joint solutions were demonstrated in the literature. 
In some cases, a template is obtained under the rigidity assumption, which might not always be fulfiled 
in practical applications \cite{Yu2015}. 

Apart from the main classes, methods for monocular scene flow (MSF) and hybrid NRSfM can be named. 
MSF jointly reconstructs non-rigid geometry and 3D displacement fields \cite{Mitiche2015}. 
In some cases, it relies on a known camera trajectory or proxy geometry (an initial coarse geometry estimate) \cite{Birkbeck2011}. 
In hybrid NRSfM, a scene-specific shape prior is obtained on-the-fly under non-rigidity, and the input is 
a sequence of point tracks \cite{Golyanik2017}. Geometry estimation is then conditioned upon the shape prior. 

MNR has only recently entered the realm of dense reconstructions \cite{Birkbeck2011, Russell2012, Yu2015}. 
The dense setting brings additional challenges for augmented reality applications such as scalability with 
the number of points and increased computational and memory complexity. 

\subsection{Contributions}

The scope of this paper is general-purpose MNR, \ie the reconstruction scenarios are not known in advance. 
We propose deep neural network (DNN) based deformation model for MNR. 
We train DNN with a new synthetically generated dataset
covering the variety of smooth and isometric deformations occurring in the real world 
(\eg clothes deformations, waving flags, bending paper and, to some extent, biological soft tissues). 
The proposed DNN architecture combines supervised learning with domain-specific loss functions. 
Our approach with a learned deformation model --- Hybrid Deformation Model Network (HDM-Net) --- surpasses performances 
of the evaluated state-of-the-art NRSfM and template-based methods by a considerable margin. We do not require dense point tracks or 
a well-positioned template. Our initialisation-free solution supports large deformations and copes well with several textures and illuminations. 
At the same time, it is robust to self-occlusions and noise. In contrast to existing DNN architectures for 3D,
we directly regress 3D point clouds (surfaces) and depart from depth maps or volumetric representations. 

In the context of MNR methods, our solution can be seen as a TBR with considerably relaxed initial conditions and 
a broader applicability range per single learned deformation model. Thus, it constitutes a new class of methods --- instead of a template, 
we rather work with a weak shape prior and a shape at rest for a known scenario class. 

We generate a new dataset which fills a gap for training DNNs for non-rigid scenes\footnote{the dataset is 
available upon request.} and perform series 
of extensive tests and comparisons with state-of-the-art MNR methods. Fig.~\ref{06_data_set_and_comparison} 
provides an overview of the proposed approach --- after training the network, we accurately infer 3D geometry 
of a deforming surface. Fig.~\ref{ARCHITECTURE} provides a high-level overview of the proposed architecture. 

The rest of the paper is partitioned in
Related Work (Sec.~\ref{sec:related_work}), 
Architecture of HDM-Net (Sec.~\ref{sec:architecture}), 
Geometry Regression and Comparisons (Sec.~\ref{sec:regressions_comparisons}) 
and Concluding Remarks (Sec.~\ref{sec:concluding_remarks}) Sections.

\section{Related Work}
\label{sec:related_work} 

In this section, we review several algorithm classes and position the proposed HDM-Net among them. 

\subsection{Non-Rigid Structure from Motion}
NRSfM requires coordinates of tracked points throughout an image sequence. 
The seminal work of Bregler \etal \cite{Bregler2000} marks the origin of batch NRSfM. 
It constrained surfaces to lie in a linear subspace of several unknown 
basis shapes. This idea was pursued by several successor methods \cite{Brand2005, Torresani2008, Paladini2012}. 
Since the basis shapes, as well as their number, are unknown, this subclass is sensitive to
noise and parameter choice. Furthermore, an optimal number of basis shapes allowing to express all 
observed deformation modes does not necessarily always exist \cite{Torresani2008}. 
Along with that, multiple further priors were proposed for NRSfM including temporal smoothness \cite{Zhu2010, Gotardo2011}, 
basis \cite{Xiao2006}, inextensibility \cite{Fayad2010, Vicente2012, Chhatkuli2018} and 
shape prior \cite{DelBue2008, Tao2013, Golyanik2017}, among others. 
The inextensibility constraint penalises deviations from configurations increasing the total surface area. 
In other words, non-dilatable states are preferred. Several methods investigate a dual trajectory basis and
considerably reduce the number of unknowns \cite{Akhter2011}, whereas the other ones explicitly
model deformations using physical laws \cite{Agudo2016PAMI}. 
Multiple general-purpose unsupervised learning techniques were successfully 
applied to NRSfM including non-linear dimensionality reduction \cite{Tao2013} (diffusion maps), 
\cite{Gotardo2011, Hamsici2012} (kernel trick) and expectation-maximisation \cite{Lee2017, Agudo_etal_cviu2018}. 
A milestone in NRSfM was accompanied by a further decrease in the number of unknowns and required prior 
knowledge for reconstruction. Thus, some of the methods perform a low-rank approximation of a stacked 
shape matrix \cite{Dai2014, Garg2013}. A further milestone is associated with the ability to perform dense reconstructions 
\cite{Garg2013, Golyanik2017, Golyanik2017d, Ansari2017, Agudo_etal_cviu2018}.

Several methods allow sequential processing \cite{Paladini2010, Zhu2010, Agudo2014}. Starting from an initial estimate 
obtained on several first frames of a sequence, they perform reconstructions upon arrival of every new frame 
in an incremental manner. The accuracy of sequential methods is consistently lower than those of the batch counterparts. 
While still relying on point tracks, they can enable lowest latencies in real-time and interactive applications. 
Several methods learn and update an elastic model of the observed scene on-the-fly \cite{Agudo2018} 
(similarly to the sequential methods, point tracks over the complete sequence are not required). 
Solving the underlying equations might be slow, and the solution was demonstrated only for sparse settings.

\begin{sidewaysfigure}
    \centering
  \includegraphics[width=1.0\linewidth]{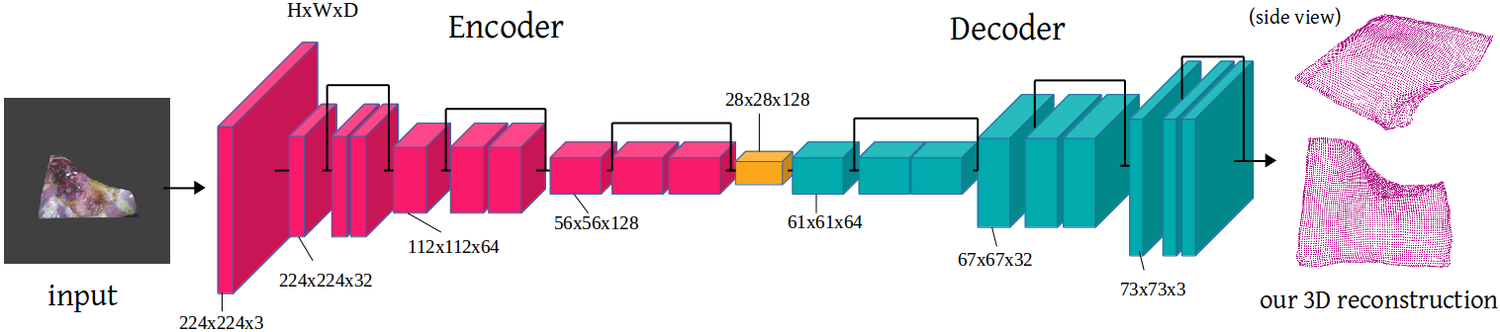} 
  \caption{An overview of the architecture of the proposed HDM-Net with encoder and decoder. The input of HDM-Net is an image of 
  dimensions $224 \times 224$ with three channels, and the output is a dense reconstructed 3D surface 
  of dimensions $73 \times 73 \times 3$ (a point cloud with $73^2$ points). } %
  \label{ARCHITECTURE} 
      \vspace{25pt}
    \centering
  \includegraphics[width=1.0\linewidth]{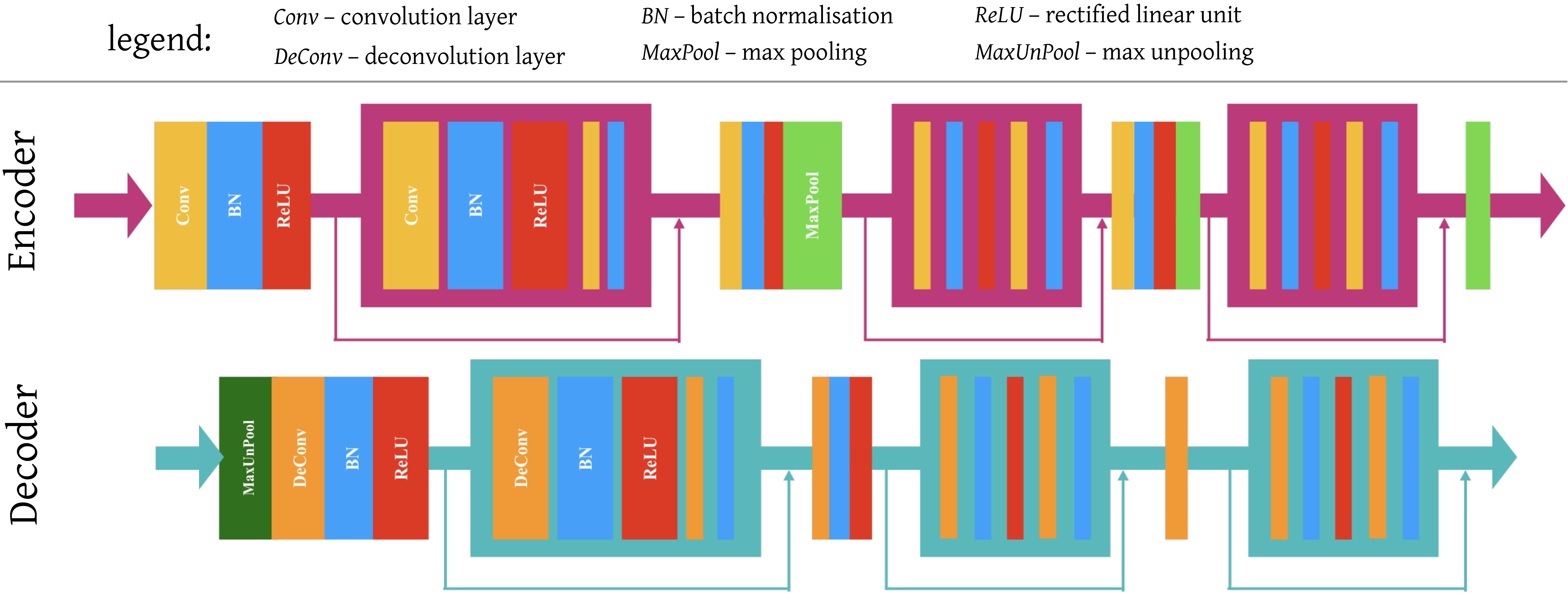} 
  \caption{Architecture of the proposed HDM-Net: detailed clarification about the structures of the encoder and decoder. }
  \label{imgpsh_fullsize2}

\end{sidewaysfigure}

\subsection{Template-Based Reconstruction}

Approaches of this class assume a known template, 
\ie an accurate reconstruction of at least one frame of the sequence. Most methods operate 
on a short window of frames or single frames. Some TBR methods are known as
non-rigid trackers \cite{Salzmann2007A}. Early physics-based techniques formalised 3D reconstruction 
with elastic models and modal analysis \cite{McInerney1993, Cohens1992}. They assumed that some material 
properties (such as the elastic modulus) of the surface are known and could handle small non-linear deformations. 

Multiple priors developed for NRSfM proved their effectiveness for TBR including isometry 
\cite{Brunet2010, Perriollat2011, Moreno-Noguer2010, Salzmann2011}, 
statistical priors \cite{Salzmann2009}, 
temporal smoothness \cite{Salzmann2007, Yu2015},  
inextensibility priors \cite{Brunet2010, Perriollat2011} and mechanical priors in an improved form \cite{Malti2013, Haouchine2014}. 
Moreover, modelling image formation process by decomposing observed intensities into lighting, shading and albedo components 
was also shown to improve tracking accuracy \cite{White2006, Moreno-Noguer2010, Gallardo2016, Liu-Yin2016, Gallardo_2017}. 

\subsection{Monocular Scene Flow}
A somewhat exotic class of approaches developed in parallel to NRSfM and TBR is monocular scene flow (MSF). 
Birkbeck \etal's approach can handle non-rigid scenes relying on a known constant camera motion \cite{Birkbeck2011}. 
While camera trajectory can be sometimes available in AR systems, there is no guarantee of its linearity. 
In \cite{Mitiche2015}, a variational solution to rigid multi-body scenes was proposed. Recently, Xiao \etal proposed 
an energy-based method for rigid MSF in the context of automotive scenarios. Their approach is based on 
a temporal velocity constancy constraint \cite{Xiao2017}. 

In general, MSF methods are restricted in the handling of non-rigid surfaces. One exception 
--- NRSfM-Flow of Golyanik \etal \cite{Golyanik2016NRSfMFlow} --- 
takes advantage of known 2D-3D correspondences and relies 
on batch NRSfM techniques for an accurate scene flow estimation of non-rigid scenes. 
It inherits the properties of NRSfM and does not assume a known camera trajectory or proxy geometry. 

\subsection{Specialised Models for Faces and Bodies}

For completeness, we provide a concise overview of specialised approaches. 
Compared to TBR, they are dedicated to the reconstruction of single object classes like 
human faces \cite{Blanz1999, Suwajanakorn2014, Garrido2016, Sela_2017} or 
human bodies \cite{Guan2009, Wandt2016}. They do not use a single prior state (a template), 
but a whole space of states with feasible deformations and variations. The models et al.
are learned from extensive data collections showing a wide variety of forms, expressions (poses) and textures. 
In almost all cases, reconstruction with these methods means projection into the space of 
known shapes.
To obtain accurate results, post-processing steps are required (\eg for transferring subtle details to 
the initial coarse estimates). In many applications, solutions with predefined models might be a right choice, 
and their accuracy and speed may be sufficient.

\subsection{DNN-based 3D Reconstruction}
In the recent three years, several promising approaches for inferring 2.5D and 3D geometry have been developed. 
Most of them regress depth maps \cite{NIPS2014, Liu2015, Garg2016, Godard2017, Tateno2017} 
or use volumetric representations \cite{choy2016, Riegler2017} akin to sign distance fields \cite{Curless1996}. 
Currently, the balance of DNN-based methods for 3D reconstruction is perhaps in favour of 
face regressors \cite{Sela_2017, Dou2017, Tewari2017, Jackson2017}. 
The alternatives to sparse NRSfM of Tome \etal and Zhou \etal work exclusively for human poses \cite{Tome2017, Zhou2018}. 
The 3D-R2N2 network generates 3D reconstructions from single and multiple views and requires large 
data collections for training \cite{choy2016}. 
In contrast to several other methods, it does not require image annotations. 
Point set generation netwoet al.rk of Fan \etal \cite{Fan_2017} is trained for a single view reconstruction of 
rigid objects and directly outputs point sets. 
More and more methods combine supervised learning and model-based losses thus imposing
additional problem-specific constraints \cite{Garg2016, Tewari2017, Fan_2017}. 
Also, this has often the side effect of decreasing the volume requirements on the datasets \cite{Godard2017, Tewari2017}. 
The work of Pumarola \etal \cite{pumarola2018geometry} is most closely related to ours. The architecture is separated into three sub-networks which have different roles --- creating heat-map of 2D images, depth estimation and 3D geometry inference. Those sub-networks are jointly trained. Our architecture is relatively simple. Encoder and decoder are employed and the output is penalized with three kinds of losses which have different geometrical properties --- 3D geometry, smooth surface and contour information after projection onto a 2D plane.

\subsection{Attributes of HDM-Net}

In this section, we position the proposed approach among the vast body of the literature on MNR. 
HDM-Net bears a resemblance to DNN-based regressors which use encoder-decoder architecture \cite{Tewari2017}. 
In contrast to many DNN-based 3D regressors \cite{choy2016, Tewari2017, Fan_2017}, our network does not include fully connected layers 
as they impede generalisability (lead to overfitting) as applied to MNR. As most 3D reconstruction approaches, 
it contains a 3D loss. 

In many cases, isometry is an effective and realistic constraint for TBR, as shown in \cite{Chhatkuli2018, Perriollat2011}. 
In HDM-Net, isometry is imposed through training data. 
The network learns the notion of isometry from the opposite, \ie by not observing other deformation modes. 
Another strong constraint in TBR is contour information which, however, has not found wide use in MSR, 
with only a few exceptions \cite{Gumerov2004, Varol2012}. In HDM-Net, we explicitly impose contour constraints 
by comparing projections of the learned and ground truth surfaces.

Under isometry, the solution space for a given contour is much better constrained compared to the extensible cases. 
The combined isometry and contour cues enable efficient occlusion handling in HDM-Net. 
Moreover, contours enable texture invariance up to a certain degree, as a contour remains unchanged irrespective of the texture. 
Next, through variation of light source positions, we train the network for the notion of shading. 
Since for every light source configuration, the underlying geometry is the same, HDM-Net
acquires awareness of varying illumination.
Besides, contours and shading in combination enable reconstruction of texture-less surfaces. 
To summarise, our framework has unique properties among MSR methods
which are rarely found in other MNR techniques, especially when combined. 

\begin{figure*}[t!]
 \centering
  \includegraphics[width=1.0\linewidth]{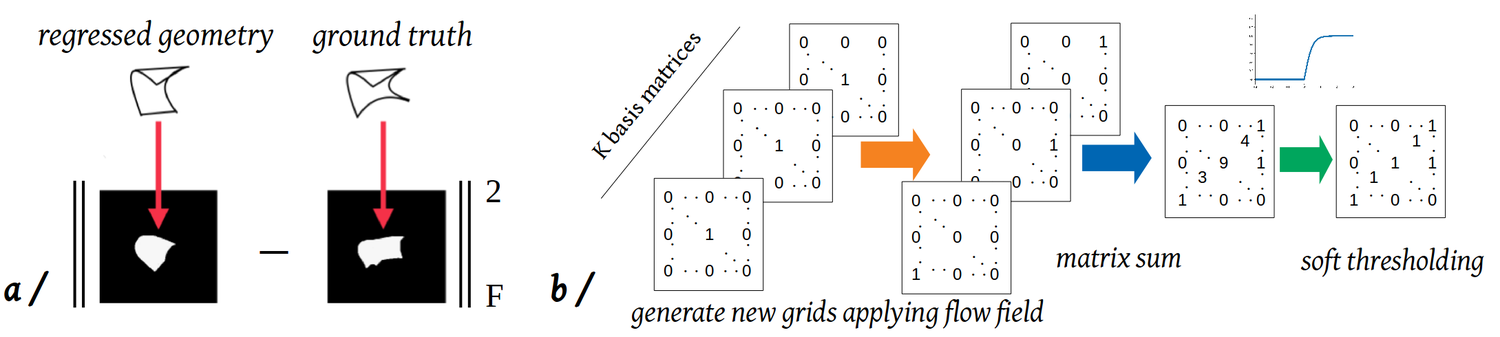} 
  \caption{Our contour loss penalises deviations between reprojections of the regressed geometry and reprojections of the ground truth. } 
  \label{05_contours} 
\end{figure*} 

\section{Architecture of HDM-Net}\label{sec:architecture} 

We propose a DNN architecture with encoder and decoder depicted in Fig.~\ref{ARCHITECTURE} (a general overview). 
The network takes as an input an image of dimensions $224 \times 224$ with three channels. 
Initially, the encoder extracts contour, shading and texture deformation cues and generates a compact latent space representation of 
dimensions \hbox{$28 \times 28 \times 128$}. Next, the decoder applies a series of deconvolutions and outputs a 3D surface 
of dimensions $73 \times 73 \times 3$ (a point cloud). It lifts the dimensionality of the latent space until the dimensionality 
of activation becomes identical to the dimensionality of ground truth.  The transition from the implicit representation into 3D 
occurs on the later stage of decoder through a deconvolution. Fig.~\ref{imgpsh_fullsize2} provides a detailed clarification about 
the structures of encoder and decoder. 

As can be seen in Fig.~\ref{ARCHITECTURE} and \ref{imgpsh_fullsize2}, we skip some connections in HDM-Net 
to avoid vanishing gradients, similar to \textit{resnet} \cite{He2016}. Due to the nature of convolutions, 
our deep network might potentially lose some important information in the forward path which might be advantageous 
in the deeper layers. 
Thus, connection skipping compensates for this side effect --- for each convolution layer --- which results 
in the increased performance. Moreover, in the backward path, shortcut connections help to overcome 
the vanishing gradient problem, \ie a series of numerically unstable gradient multiplications leading 
to vanishing gradients. Thus, the gradients are successfully passed to the shallow layers. 

Fully connected (FC) layers are often used in classification tasks \cite{NIPS2012}. They have more
parameters than convolution layers and are known as a frequent cause of overfitting. We have tried FC layers 
in HDM-Net and observed overfitting on the training dataset. Thus, FC layers reduce generalisation ability 
of our network. Furthermore, spatial information is destroyed as the data in the decoder is concatenated
before being passed to the FC layer. In our task, needless to say, spatial cues are essential for 3D regression. 
In the end, we omit FC layers and successfully show generalisation ability of 3D reconstruction on the 
test data.

\subsection{Loss Functions}

Let $\bfS = \{\bfS_f\}$, $f \in \{1, \hdots, F\}$ denote predicted 3D states, and $\bfS^{GT} = \{\bfS_{f}^{GT}\}$ is 
the ground truth geometry; 
$F$ is the total number of frames and $N$ is the number of points in the 3D surface. 
In HDM-Net, contour similarity and the isometry constraint are the key innovations
and we apply three types of loss functions summarised into the loss energy: 
\begin{equation}
  \bfE(\bfS, \bfS^{GT}) = \bfE_{3D}(\bfS, \bfS^{GT}) + \bfE_{iso}(\bfS) + \bfE_{cont.}(\bfS, \bfS^{GT}). 
\end{equation}

\noindent\textbf{3D error:} The 3D loss is the main loss in 3D regression. 
It penalises the differences between predicted and ground truth 3D states and is 
common in training for 3D data: 
\begin{equation}\label{eq:loss_3D}
\bfE_{3D}(\bfS, \bfS^{GT})=\frac{1}{F}\sum_{f=1}^{F} \lVert \bfS_{f}^{GT} - \bfS_{f}\rVert_\mathcal{F}^{2}, 
\end{equation}
where $\lVert \cdot \rVert_{\mathcal{F}}$ denotes the Frobenius norm. 
Note that we take an average of the squared Frobenius norms of the differences between the learned and ground truth geometries. 

\noindent\textbf{Isometry prior:} To additionally constrain the regression space, we embed 
isometry loss which enforces the neighbouring vertices to be located close to each other. 
Several versions of inextensibility and isometry constraints can be found in MSR --- a common one is based on 
differences between Euclidean and geodesic distances. For our DNN architecture, we choose a differentiable loss 
which performs Gaussian smoothing of $\bfS_{f}$ and penalises the difference between the unembellished and smoothed %
version $\hat{\bfS}_i$: 
\begin{equation}
\bfE_{iso}(\bfS) = \frac{1}{F} \sum_{f=1}^{F} \lVert \hat{\bfS}_f - \bfS_{f} \rVert_\mathcal{F}, 
\end{equation}
with 
\begin{equation}
\hat{\bfS}_f=\frac{1}{2\pi\sigma^{2}} \operatorname{exp}\left ( -\frac{x^{2}+y^{2}}{2\sigma^{2}}\right)\ast \bfS_f, 
\end{equation}
where $\ast$ denotes a convolution operator and $\sigma^2$ is the variance of Gaussian. 

\begin{figure*}[t!]
 \centering
  \includegraphics[width=1.0\linewidth]{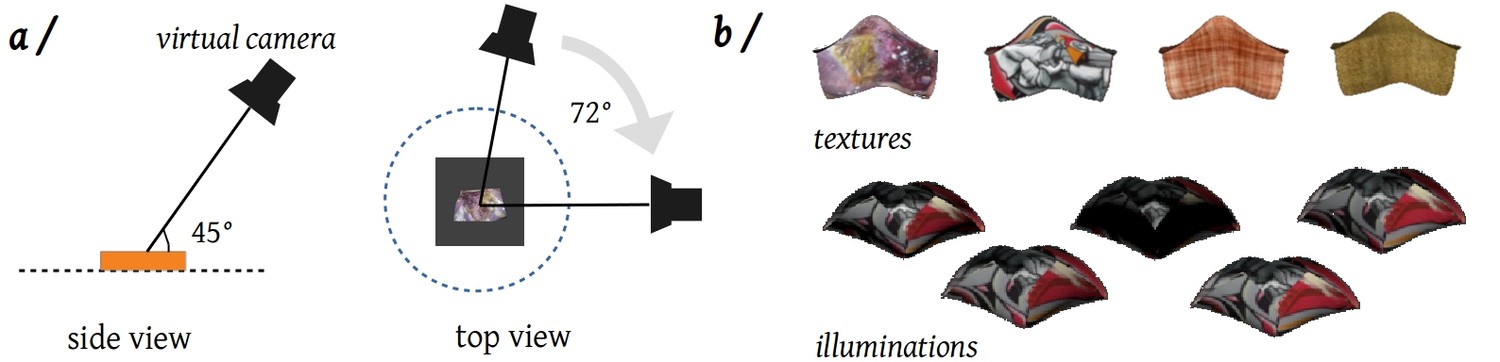} 
  \caption{ Camera poses used for the dataset generation (a); different textures applied to the dataset: \textit{endoscopy}, 
  \textit{graffiti}, \textit{clothes} and \textit{carpet} (b-top) and different illuminations (b-bottom). } 
  \label{07_textures} 
\end{figure*}

\noindent\textbf{Contour loss:} If the output of the network and the ground truth coordinates are similar, 
the contour shapes after projection onto a 2D plane have to be similar as well. The main idea of the reprojection loss 
is visualised in Fig.~\ref{05_contours}-(a). 
After the inference of the 3D coordinates by the network, we project them onto the 2D plane and compute the difference 
between the two projected contours. 
If focal lengths $f_x$, $f_y$ as well as the principal point $(c_x, c_y)$ of the camera are known 
(the $\bfK$ used for the dataset generation is provided in Sec.~\ref{sec:data_sets_and_training}), 
observed 3D points $\bfp = (p_x, p_y, p_z)$ are projected to the 
image plane by the projection operator $\pi$ : $\mathbb{R}^3 \rightarrow \mathbb{R}^2$: 
\begin{align}\label{eq:projection}
  \bfp'(u, v) = \pi(\bfp) = \Bigg( f_x \frac{p_x}{p_z} + c_x, f_y \frac{p_y}{p_z} + c_y \Bigg)^\mathsf{T},  
\end{align} 
where $\bfp'$ is the 2D projection of $\bfp$ with 2D coordinates $u$ and $v$. Otherwise, we apply an orthographic camera model. 

A na\"ive shadow casting of a 3D point cloud onto a 2D plane is not differentiable, \ie the network cannot backpropagate gradients 
to update the network parameters. The reason is twofold. 
In particular, the cause for indifferentiability is 
the transition from point intensities to binary shadow indicators with an ordinary step function (the numerical reason) 
using point coordinates as indexes on the image grid (the framework-related reason). 

Fig.~\ref{05_contours}-(b) shows how we circumvent this problem. 
The first step of the procedure is the projection of 3D coordinates onto a 2D plane using either a perspective 
or an orthographic projection. As a result of this step, we obtain a set of 2D points. 
We generate $K = 73^2$ translation matrices $\bfT_j$ \tiny$ = \begin{pmatrix}
                                                                1 & 0 & u \\
                                                                0 & 1 & v
                                                              \end{pmatrix}$ 
\normalsize using 2D points and a flow field tensor of dimension $K \times 99 \times 99 \times 2$
(the size of each binary image is $99 \times 99$). 
Next, we apply bilinear interpolation \cite{Jaderberg2015} with generated flow fields on the replicated basis matrix $\bfB$ $K$ times 
and obtain $K$ translation indicators. 
$\bfB_{99 \times 99}$ is a sparse matrix with only a single central non-zero element which equals to $1$. 
Finally, we sum up all translation indicators and softly threshold positive values in the sums to $\approx 1$, \ie our shadow indicator. 
Note that to avoid indifferentiability in the last step, the thresholding is performed by a combination of a rectified 
linear unit (ReLU) and $\operatorname{tanh}$ function (see Fig.~\ref{05_contours}-(b)): 
\begin{equation}
  \tau( \mathcal{I}(\bfs_f(n))) = \max(\tanh(2 \, \bfS_f(n)), 0), %
\end{equation}
where $n \in \{1, \hdots, N\}$ denotes the point index, $\bfs_f(n)$ denotes a reprojected point $\bfS_f(n)$ in frame $f$, 
and $\mathcal{I}(\cdot)$ fetches intensity of a given point. We denote the differentiable projection operator and differentiable soft thresholding 
operator by the symbols $\pi^{\dagger}(\cdot)$ and $\tau(\cdot)$ respectively. 
Finally, the contour loss reads 
\begin{equation}
  \bfE_{cont.}(\bfS, \bfS^{GT}) = \frac{1}{F} \sum_{f = 1}^{F} \lVert \tau\big(\pi^{\dagger}\big(\bfS_f\big)\big) - 
  \tau \big( \pi^{\dagger} \big(\bfS_f^{GT}\big)\big) \rVert_\mathcal{F}^2. 
\end{equation}
Note that object contours correspond to $0$-$1$ transitions. 

\vspace{0.5cm}

\section{Dataset and Training}\label{sec:data_sets_and_training} 

For our study, we generated a dataset with a non-rigidly deforming object using \textit{Blender} \cite{blender2018}. 
In total, there are $4648$ different temporally smooth 3D deformation states with structure bendings, smooth foldings 
and wavings, rendered under Cook-Torrance illumination model \cite{Cook1982} 
(see Fig.~\ref{06_data_set_and_comparison} for the exemplary frames from our dataset). 
We have applied five different camera poses, 
five different light source positions and four different textures corresponding 
to the scenarios we are interested in --- \textit{endoscopy}, \textit{graffiti} (it resembles a waving flag)
\textit{clothes} and \textit{carpet} (an example of an arbitrary texture). 
The endoscopic texture is taken from \cite{Giannarou2013}. 
\begin{figure}
 \centering
  \includegraphics[width=1.0\linewidth]{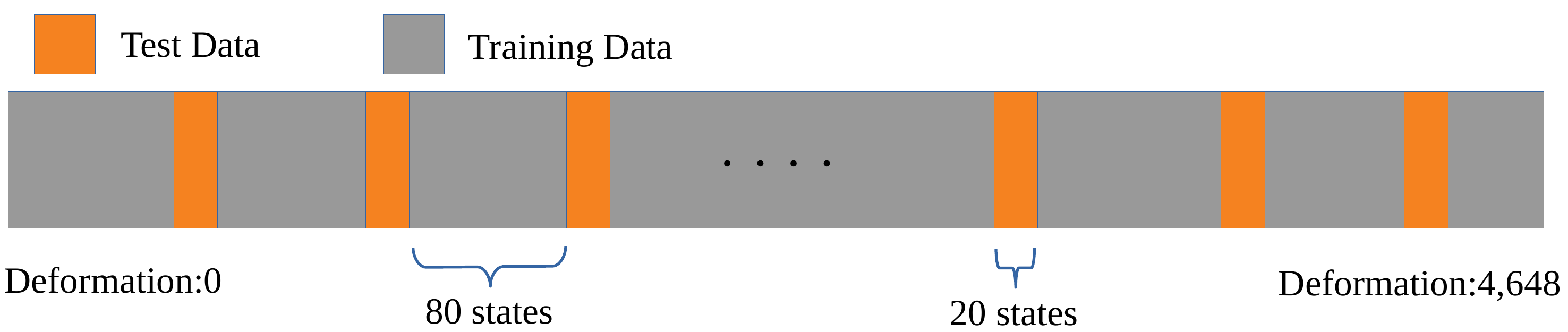} 
  \caption{The pattern of the training and test datasets. } 
  \label{04_test_and_training} 
\end{figure} 
Illuminations are generated based on the scheme in Fig.~\ref{07_textures}-(a), the textures and illuminations are shown in Fig.~\ref{07_textures}-(b). 
We project the generated 3D scene by a virtual camera onto a 2D plane upon Eq.~\eqref{eq:projection}, with \hbox{
$\bfK =$
\tiny
$\begin{pmatrix}	
         280 & 0 & 128   \\ 
         0 & 497.7 & 128 \\
         0 & 0 & 1
        \end{pmatrix}
$
\normalsize}. The background in every image is of the same opaque colour. 
We split the data into training and test subsets in a repetitive manner, see Fig.~\ref{04_test_and_training} for the pattern. 
We train HDM-Net jointly on several textures and illuminations, with the purpose of illumination-invariant and texture-invariant 
regression. One illumination and one texture are reserved for the test dataset exclusively. 
Our images are of the dimensions $256 \times 256$. They reside in $15.2$ Gb of memory, and the ground truth geometry 
requires $1.2$ Gb (in total, $16.4$ Gb). The hardware configuration consists of two six-core processors 
Intel(R) Xeon(R) CPU E5-1650 v4 running at $3.60$GHz, 16 GB RAM and a GEFORCE GTX 1080Ti GPU with 11GB of global memory. 
In total, we train for $95$ epochs, and the training takes two days in \textit{pytorch} \cite{pytorch2018, paszke2017automatic}. 
The evolution of the loss energy is visualised in Fig.~\ref{12_graphs}-(a). 
The inference of one state takes ca.~$5$ ms.

\vspace{0.5cm}
\section{Geometry Regression and Comparisons}
\label{sec:regressions_comparisons}

\begin{sidewaysfigure}
 \centering 
  \includegraphics[width=22cm,height=11cm]{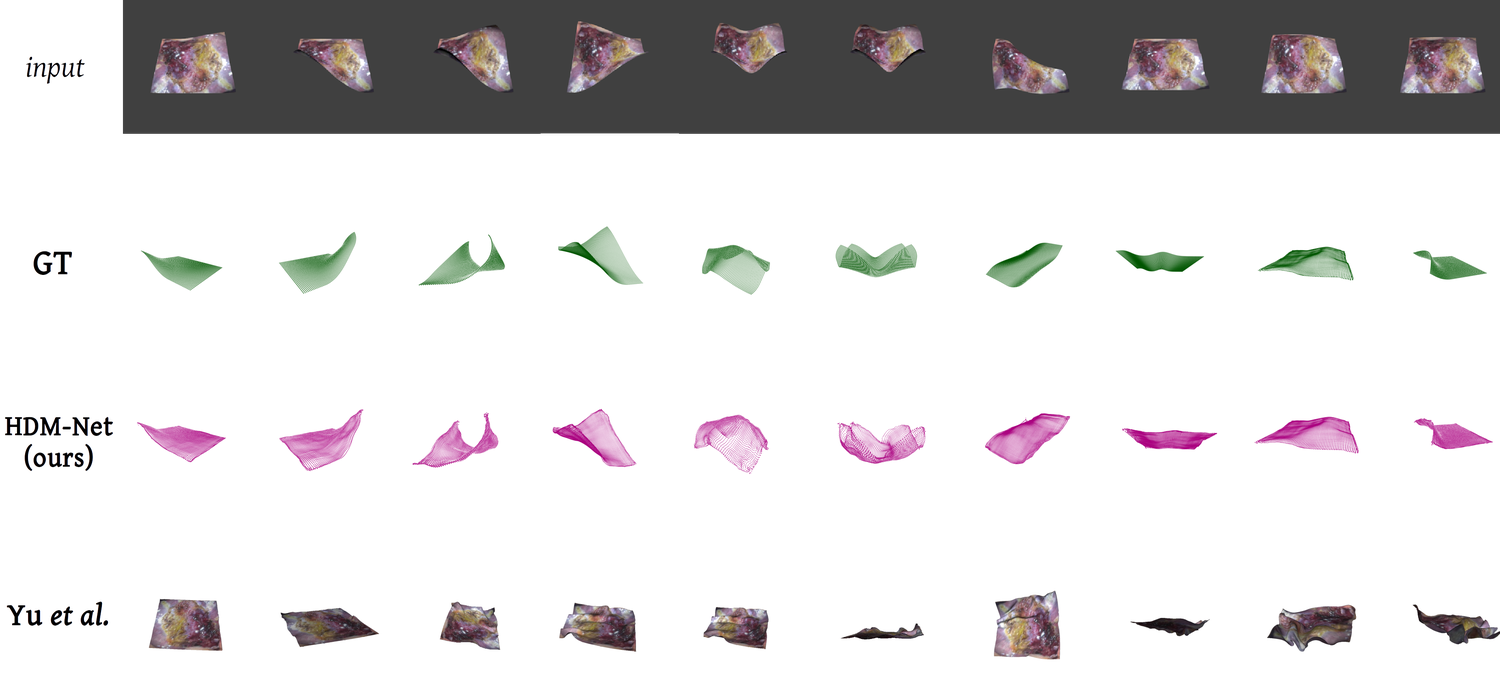} 
  \caption{ Selected reconstruction results on endoscopically textured surfaces for HDM-Net (our method) and Yu \etal \cite{Yu2015}. } 
  \label{10_} 
\end{sidewaysfigure}

We compare our method with the template-based reconstruction of Yu \etal \cite{Yu2015}, variational NRSfM approach (VA) of Garg \etal \cite{Garg2013} 
and NRSfM method of Golyanik \etal \cite{Golyanik2017d} --- Accelerated Metric Projections (AMP). 
We use an optimised heterogeneous CPU-GPU version of VA written in C++ and CUDA C \cite{cuda2018}. 
AMP is a C++ CPU version which relies on an efficient solution of a semi-definite programming 
problem and is currently one of the fastest batch NRSfM methods. 
For VA and AMP, we compute required dense point tracks.
Following the standard praxis in NRSfM, we project the ground truth shapes onto a virtual 
image plane by a slowly moving virtual camera. Camera rotations are parametrised 
by Euler angles around the $x$-, $y$- and $z$-axes. 
We rotate for up to $20$ degrees around each axis, with five degrees per frame.
This variety in motion yields minimal depth changes required for an accurate initialisation in NRSfM. 
We report runtimes, 3D error
\begin{equation}
e_{3D} = \frac{1}{F} \sum_{f = 1}^{F} \frac{\lVert \bfS_{f}^{GT} - \bfS_{f} \rVert_{\mathcal{F}}}{\lVert \bfS_{f}^{GT} 
\rVert_{\mathcal{F}}}  
\end{equation}
and standard deviation $\sigma$ of $e_{3D}$. 
Before computing $e_{3D}$, we align $\bfS_f$ and the corresponding $\bfS_f^{GT}$ with Procrustes analysis. 

Runtimes, $e_{3D}$ and $\sigma$ for all three methods are summarised in Table~\ref{runtimes}. 
AMP achieves around $30$ \textit{fps} and can execute only for $100$ frames per batch at a time. 
However, this estimate does not include often prohibitive 
computation time of dense correspondences with multi-frame optical flow methods such as \cite{Taetz2016}. 
Note that runtime of batch NRSfM depends on the batch size, and the batch size 
influences the accuracy and ability to reconstruct.
VA takes advantage of a GPU and executes with $2.5$ \textit{fps}. 
Yu \etal \cite{Yu2015} achieves around $0.3$ \textit{fps}. In contrast, HDM-Net processes one frame 
in only $5$ ms. This is by far faster than the compared methods. 
Thus, HDM-Net can compete in runtime with rigid structure from motion 
\cite{TomasiKanade92}. The runtime of the latter method is still considered as the lower runtime bound for NRSfM\footnote{when executed 
in a batch of $100$ frames with $73^2$ points each, a C++ version of \cite{TomasiKanade92} takes 
$1.47$ ms per frame on our hardware; for $400$ frames long batch, it requires $5.27$ ms per frame. }. 

At the same time, the accuracy of HDM-Net is the highest among all tested methods. 
Selected results with complex deformations are shown in Fig.~\ref{10_}. 
We see that Yu \etal \cite{Yu2015} copes well with rather small 
deformations, and our approach accurately resolves even challenging cases not exposed during the training.
In the case of Yu \etal \cite{Yu2015}, 
the high $e_{3D}$ is explained by a weak handling of self-occlusions and large deformations.
In the case of NRSfM methods, the reason for the high $e_{3D}$ is an inaccurate initialisation. Moreover, VA 
does not handle foldings and large deformations well.

\begin{table}[t!]
\parbox{.5\linewidth}{
\centering
        \scalebox{0.98}{%
\begin{tabular}{ccccc} \toprule 
				& Yu \etal \cite{Yu2015} 	& AMP \cite{Golyanik2017d}	& VA \cite{Garg2013} 	& HDM-Net  		\\ \midrule 
      $t$, \textit{s}  		& 3.305	 			& 0.035				& 0.39  		& \textbf{0.005}	\\ 	
      $e_{3D}$  		& 1.3258	 		& 1.6189			& 0.46			& \textbf{0.0251}	\\ 
      $\sigma$			&  \textbf{0.0077}			& 1.23				& 0.0334		& 0.03		\\ \hline
\end{tabular} 
        }
\caption{Per-frame runtime $t$ in \textit{seconds}, $e_{3D}$ and $\sigma$ comparisons of Yu \etal \cite{Yu2015}, AMP \cite{Golyanik2017d} and HDM-Net (proposed method).  }\label{runtimes}

        \scalebox{0.98}{
\begin{tabular}{ccccc}\toprule 
			&\textit{endoscopy} 	& \textit{graffiti} 	& \textit{clothes} 	& \textit{carpet}	\\ \midrule 
	$e_{3D}$	&\textbf{0.0485}	& 0.0499		& 0.0489		& 0.1442		\\    
	$\sigma$	& \textbf{0.01356}	& 0.022			& 0.02648		& 0.02694		\\ \hline 
 \end{tabular}
 }

\caption{Comparison of 3D error for different textures and the same illumination (number $1$). } \label{avg_errors_textures}

    }
    \hfill
    \parbox{.45\linewidth}{
        \centering
\vspace{-6pt}
        \scalebox{0.95}{

\begin{tabular}{cccccc}\toprule 
		& \textit{illum.~1} 	& \textit{illum.~2} 	& \textit{illum.~3} 	& \textit{illum.~4}	& \textit{illum.~5} 	\\ \midrule 
		\addlinespace[0.28cm]
      $e_{3D}$  & 0.07952	 	& 0.0801		& 0.07942		& \textbf{0.07845}	& \textbf{0.07827}	\\ 
      \addlinespace[0.3cm]
      $\sigma$	& \textbf{0.0525}	&  0.0742		& 0.0888		& 0.1009		& 0.1123		\\ \hline 
      \addlinespace[0.2cm]
 \end{tabular}
        }
\vspace{-10pt}
\caption{Comparison of 3D error for different illuminations. }
\label{avg_errors_illuminations}
\vspace{11pt}
\scalebox{0.88}{
\begin{tabular}{c|c|c|c|c}\toprule 
			& 3D        & 3D + Con. 	        & 3D + Iso. 	& 3D + Con. + Iso.	\\ \midrule 
	$e_{3D}$	& 0.0698 	& \textbf{0.0688}	& 0.0784		& 0.0773		\\    
	$\sigma$	& \textbf{0.0761} 	& 0.0736			& 0.0784	& 0.0789		\\ \hline 
 \end{tabular}}

\caption{Comparison of effects of loss functions. } 
\label{loss_comparison}
    }
    
\end{table}

\begin{figure} [t!]
\begin{minipage}{0.5\textwidth}
 \centering 
  \includegraphics[width=7cm, height=5cm]{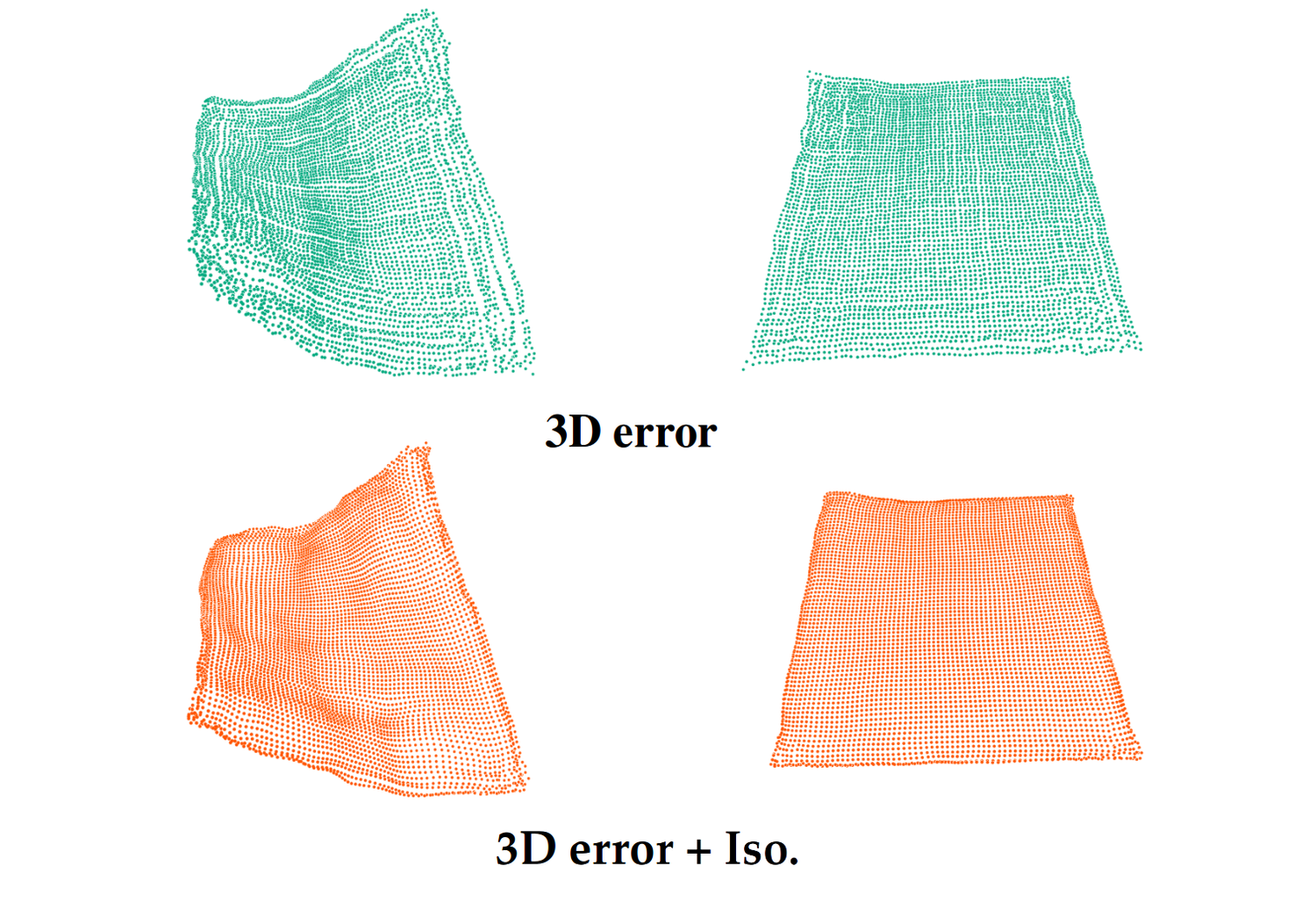} 
  \caption{Comparison of 3D reconstruction with 3D error (top row) and 3D error + isometry prior (bottom row)} 
  \label{loss_comp} 
  \end{minipage}\hfill
  \begin{minipage}{0.48\textwidth}
 \centering 
  \includegraphics[width=1.0\linewidth]{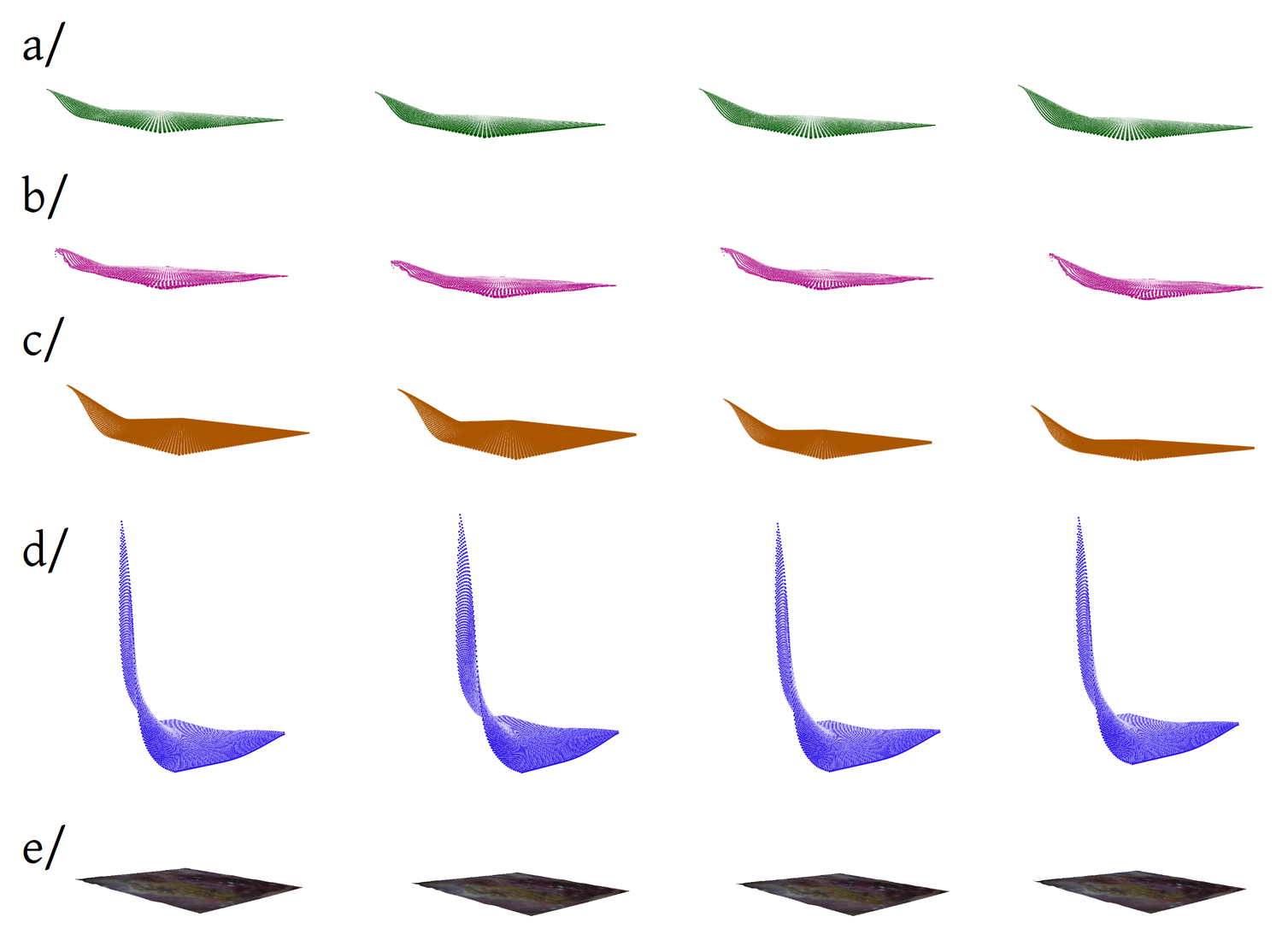} 
  \caption{ Qualitative comparisons of ground truth (a), 
  HDM-Net (proposed method) (b),
  AMP \cite{Golyanik2017d} (c),
  VA \cite{Garg2013} (d)
  and Yu \etal \cite{Yu2015} (e) on several frames of our test sequence from the first $100$ frames (each column corresponds to one frame). } 
  \label{11_second_vis} 
  \end{minipage}
\end{figure}

\begin{figure}[t!]
 \centering 
  \includegraphics[width=1.0\linewidth]{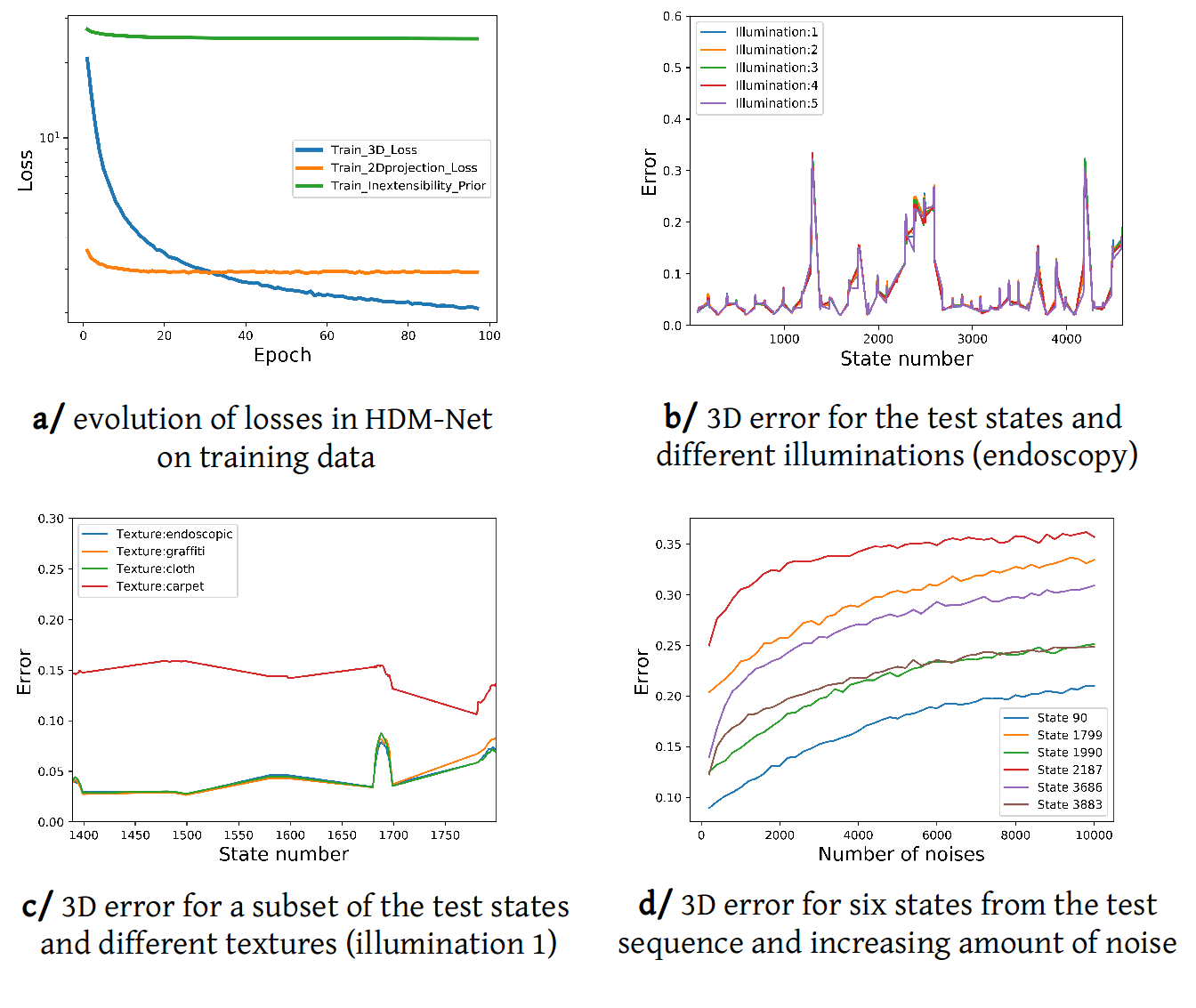} 
  \caption{ Graphs of $e_{3D}$ for varying illuminations (for \textit{endoscopy} texture), varying textures 
  (for illumination 1) as well as six states under increasing amount of noise. Note that in b/ and c/, only 
  the errors obtained on the test data are plotted. For c/, HDM-Net was trained on a subset of training states (three main textures and one illumination). } 
  \label{12_graphs} 
\end{figure}

Table~\ref{avg_errors_illuminations} summarises $e_{3D}$ for our method under different illumination conditions. 
We notice that our network copes well with all generated illuminations ---
the difference in $e_{3D}$ is under $3\%$. Table~\ref{avg_errors_textures} shows $e_{3D}$ comparison for different 
textures. Here, the accuracy of HDM-Net drops on the previously unseen texture by the factor of three, which still corresponds 
to reasonable reconstructions with the captured main deformation mode. 
Another quantitative comparison is shown in Fig.~\ref{11_second_vis}. In this example, all methods execute on the first $100$ frames of the sequence. 
AMP \cite{Golyanik2017d} captures the main deformation mode with $e_{3D} = 0.1564$ but struggles to perform a fine-grained distinction 
(in Table~\ref{runtimes}, $e_{3D}$ is reported over the sequence of $400$ frames, hence the differing metrics). 
VA suffers under an inaccurate initialisation under rigidity assumption and Yu \etal \cite{Yu2015}, by contrast,
does not recognise the variations in the structure.
All in all, HDM-Net copes well with self-occlusions. Graphs of $e_{3D}$ as functions of the state index under varying illuminations and textures 
can be found in Fig.~\ref{12_graphs}-(b,c). Table~\ref{loss_comparison} shows the comparison of $e_{3D}$ using networks trained with various combinations of loss functions. \textit{3D + Con.} shows the lowest $e_{3D}$ and applying \textit{isometry prior} increases $e_{3D}$. Since \textit{isometry prior} is smoothing loss, the 3D grid becomes smaller in comparison to the outputs without \textit{isometry prior} hence higher $e_{3D}$.  However, as shown in Fig.~\ref{loss_comp}, isometry prior allows the network to generate smoother 3D geometries preserving deformation states.

\begin{figure}[t!]
 \centering 
  \includegraphics[width=1.0\linewidth]{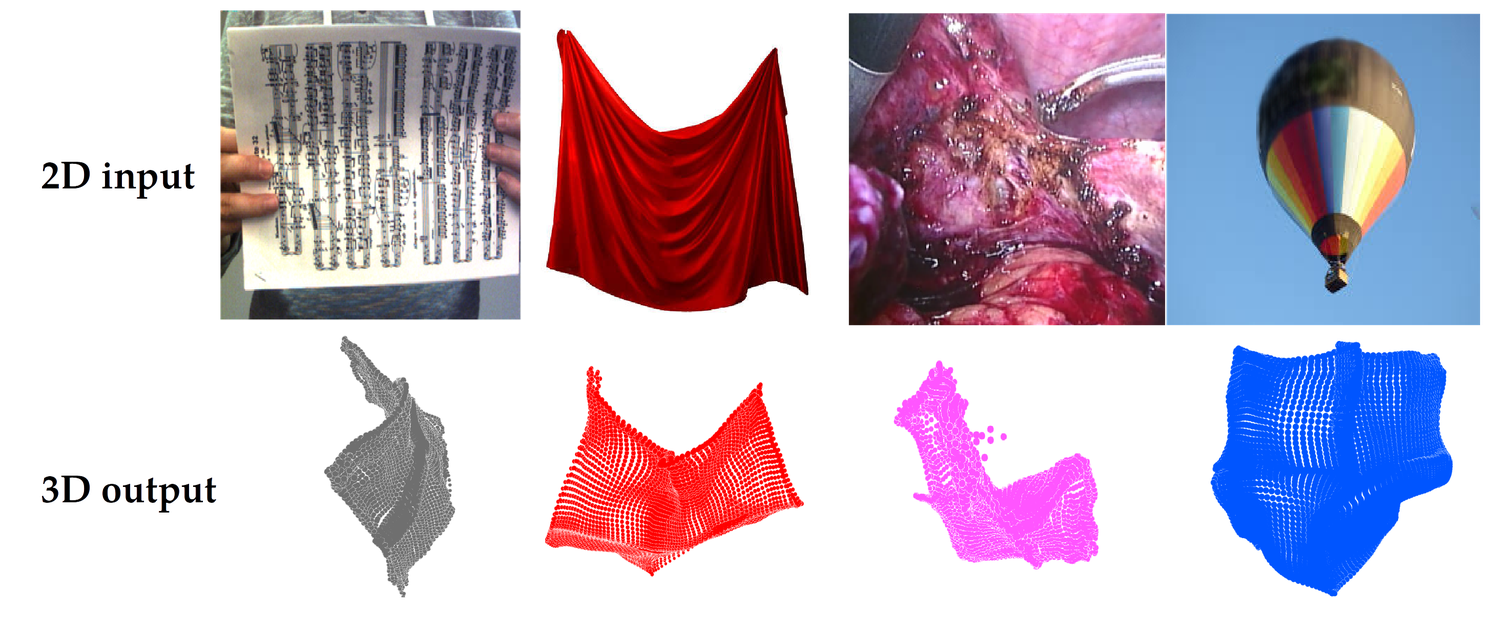} 
  \caption{Exemplary reconstructions from real images obtained by HDM-Net (music notes, a fabric, surgery and an air balloon)} 
  \label{real_world} 
\end{figure} 
Next, we evaluate the performance of HDM-Net on noisy input images. Therefore, we augment the dataset with increasing amounts of uniform 
salt-pepper noise. Fig.~\ref{12_graphs}-(d) shows the evolution of the $e_{3D}$ as a function of the amount of noise, 
for several exemplary frames corresponding to different input difficulties for the network. We observe that HDM-Net is well-posed \hbox{w.r.t} 
noise --- starting from the respective values obtained for the noiseless images, the $e_{3D}$ increases gradually.

We tested HDM-Net on several challenging real images. 
Fig.~\ref{real_world} shows the tested images and our reconstructions. We recorded a music note image for an evaluation of our network in real-world scenario.
Despite different origin of the inputs (music notes, a fabric \cite{WrincklesHanging0037}, an endoscopic view during a surgery \cite{Giannarou2013} and an air balloon \cite{balloon}), HDM-Net produces realistic and plausible results.
Note how different are the regressed geometries which suggests the generalisation ability of the proposed solution. 

In many real-world cases, HDM-Net produces acceptable results. 
However, if the observed states differ a lot from the states in the training data, HDM-Net might fail to recognise and regress the state. This can be addressed by an extension or tailoring of the data set for specific cases. Adding training data originating 
from motion and geometry capture of real objects might also be an option.

\section{Concluding Remarks}\label{sec:concluding_remarks} 

We have presented a new monocular surface recovery method with a deformation model replaced by a DNN --- HDM-Net. 
The new method reconstructs time-varying geometry from a single image and is robust to self-occlusions, changing 
illumination and varying texture. Our DNN architecture consists of an encoder, a latent space and a decoder, and is furnished with 
three domain-specific losses. Apart from the conventional 3D data loss, we propose isometry and reprojection losses. 
We train HDM-Net with a newly generated dataset with ca.~four an a half thousands states, four different illuminations, five different camera poses 
and three different textures. Experimental results show the validity of our approach and its suitability for 
reconstruction of small and moderate isometric deformations under self-occlusions. Comparisons with one template-based and two template-free 
methods have demonstrated a higher accuracy in favour of HDM-Net. 
Since HDM-Net is one of the first approach of the new kind, there are multiple avenues for investigations and improvements. One apparent
direction is the further augmentation of the test dataset with different backgrounds, textures and illuminations. Next, 
we are going to test more advanced architectures such as generative adversarial networks and %
recurrent connections for the enhanced temporal smoothness. Currently, we are also investigating the relevance of HDM-Net for medical applications 
with augmentation of soft biological tissues.

\section*{Acknowledgement}
Development of HDM-Net was supported by the project DYMANICS (01IW15003) of the German Federal Ministry of Education and Research (BMBF). The authors thank NVIDIA Corporation for the hardware donations.

\vspace{0.35cm}

\bibliographystyle{splncs04}
\bibliography{paper}
\end{document}